\documentclass[sigconf]{acmart}

\usepackage{balance}
  
\usepackage{booktabs} 
\usepackage{graphicx}
\usepackage{amsmath,amssymb} 
\usepackage{makecell}
\usepackage{multirow}
\usepackage{subfigure}

\newcommand{\etal}{\textit{et al}. }
\newcommand{\ie}{\textit{i.e.} }
\newcommand{\eg}{\textit{e.g.} }

\newcommand{\fig}{Figure }
\newcommand{\tab}{Table }
\newcommand{\eqn}{Eq. }

\def\Mat#1{{\boldsymbol{#1}}}

\graphicspath{{./figures}}  

\def\BibTeX{{\rm B\kern-.05em{\sc i\kern-.025em b}\kern-.08emT\kern-.1667em\lower.7ex\hbox{E}\kern-.125emX}}
    
\copyrightyear{2019} 
\acmYear{2019} 
\acmConference[MM '19]{Proceedings of the 27th ACM International Conference on Multimedia}{October 21--25, 2019}{Nice, France}
\acmBooktitle{Proceedings of the 27th ACM International Conference on Multimedia (MM '19), October 21--25, 2019, Nice, France}
\acmPrice{15.00}
\acmDOI{10.1145/3343031.3351040}
\acmISBN{978-1-4503-6889-6/19/10}

\begin{document}

\fancyhead{}

\title{Explainable Video Action Reasoning via Prior Knowledge and State Transitions}



\author{Tao Zhuo}
\affiliation{\institution{National University of Singapore}}
\email{zhuotao@nus.edu.sg}
	
\author{Zhiyong Cheng}
\affiliation{\institution{Shandong Computer Center (National Supercomputer Center in Jinan), Qilu University of Technology (Shandong Academy of Sciences)}}
\email{jason.zy.cheng@gmail.com}

\author{Peng Zhang}
\affiliation{\institution{Northwestern Polytechnical University}}
\email{zh0036ng@nwpu.edu.cn}

\author{Yongkang Wong}
\affiliation{\institution{National University of Singapore}}
\email{yongkang.wong@nus.edu.sg}

\author{Mohan Kankanhalli}
\affiliation{\institution{National University of Singapore}}
\email{mohan@comp.nus.edu.sg}

%

\begin{abstract}

Human action analysis and understanding in videos is an important and challenging task. Although substantial progress has been made in past years, the explainability of existing methods is still limited. In this work, we propose a novel action reasoning framework that uses prior knowledge to explain semantic-level observations of video state changes. Our method takes advantage of both classical reasoning and modern deep learning approaches. Specifically, prior knowledge is defined as the information of a target video domain, including a set of objects, attributes and relationships in the target video domain, as well as relevant actions defined by the temporal attribute and relationship changes (\ie state transitions). Given a video sequence, we first generate a scene graph on each frame to represent concerned objects, attributes and relationships. Then those scene graphs are linked by tracking objects across frames to form a spatio-temporal graph (also called video graph), which represents semantic-level video states. Finally, by sequentially examining each state transition in the video graph, our method can detect and explain how those actions are executed with prior knowledge, just like the logical manner of thinking by humans. Compared to previous works, the action 
reasoning results of our method can be explained by both logical rules and semantic-level observations of video content changes. Besides, the proposed method can be used to detect multiple concurrent actions with detailed information, such as \emph{who} (particular objects), \emph{when} (time), \emph{where} (object locations) and \emph{how} (what kind of changes). Experiments on a re-annotated dataset CAD-120 show the effectiveness of our method.
	
\end{abstract}

%
%


%


%
\maketitle

\section{Introduction}
\begin{figure}[t]
	\centering
	\includegraphics[width=0.5\textwidth]{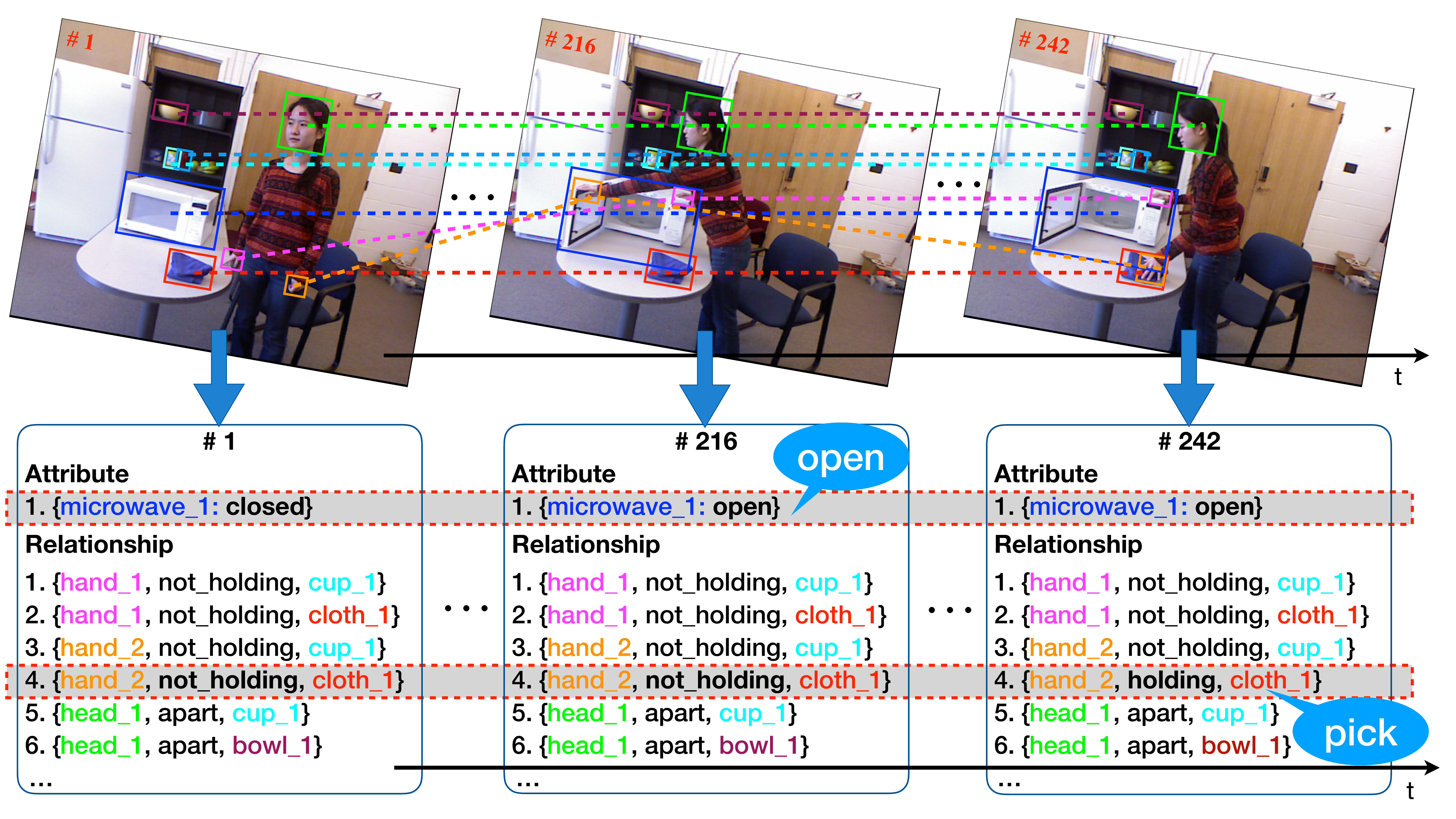}
	\caption{Video action reasoning with prior knowledge and semantic-level state transitions.
		All of the concerned objects, attributes and relationships are represented in the generated video graph, each attribute and relationship transition can be explained as a performed action by logical rules. Two actions ``open'' (\emph{microwave\_1}, \emph{closed} to \emph{open}, frame 216) and ``pick'' (\emph{hand\_2} and \emph{cloth\_1}, \emph{not\_holding} to \emph{holding}, frame 242) can be detected and explained by the attribute and relationship changes, respectively.}
	\label{fig_video_graph}
\end{figure}

Human action analysis and understanding in videos is an important problem of multimedia content analysis and a crucial component of human-machine interaction systems. Recently, with the success of deep learning in a variety of computer vision tasks, great progress has been achieved in video action recognition with various deep neural networks~\cite{CVPR2017_Carreira,CVPR2018_Wang,CVPR2018_Feichtenhofer,CVPR2018_Tran,TCSVT2018_Wang}. Compared to early action recognition approaches~\cite{CVIU2001_Intille,ECCV2008_Tran,AAAI2010_Ijsselmuiden,CVPR2011_Morariu,CVPR2011_Brendel} that perform logical reasoning by using rules on low-level features (\eg gradients, motion, location and trajectories), deep learning based methods can take advantage of the semantic-level information (\eg attributes of an object and relationship between objects) of video frames. However, due to the lack of rules for logical reasoning, most of the existing deep learning based action recognition methods~\cite{CVPR2017_Carreira,CVPR2018_Wang,CVPR2018_Feichtenhofer,CVPR2018_Tran} cannot provide detailed information to explain how an action is performed, such as \emph{who} (particular objects), \emph{when} (time), \emph{where} (object locations) and \emph{how} (what kind of changes). In this paper, we develop a novel video action reasoning framework that uses rules to understand and explain the semantic-level video state changes, which effectively bridges the gap between the classical reasoning and the modern deep learning based approaches.

Deep neural networks have been widely used in video action recognition from various perspectives. The popular two-stream convolutional networks~\cite{CVPR2017_Carreira,CVPR2018_Wang,CVPR2018_Feichtenhofer,CVPR2018_Tran} can capture the complementary information on appearance from still frames and motion between frames. Besides, spatio-temporal graphs with Recurrent Neural Networks (RNN)~\cite{CVPR2016_Jain} or Graph Convolutional Networks (GCN)~\cite{ICLR2017_Kipf,ECCV2018_Wang,ECCV2018_Guo,ECCV2018_Qi} focus on the structured video representation.
Recently, with the advances of deep learning in scene graph representation~\cite{IJCV2017_Krishna,CVPR2017_Dai,CVPR2017_Xu}, researchers attempt to use attributes of an object and the relationship between objects for semantic-level video content understanding. For example, Alayrac \etal~\cite{ICCV2017_Alayrac} automatically discovered the states of objects (\eg \emph{empty/full} of a cup) and associated manipulation actions. Liu \etal~\cite{ICCV2017_Liu} jointly recognized object fluents (\ie, changeable attributes of objects and relationships between objects) and tasks in egocentric videos. Zhu \etal~\cite{ICCV2017_Zhu} predicted a sequence of actions from visual semantic observations. Action recognition accuracy has significantly improved in the past few years. However, without rules for logical reasoning, the explainability of those deep learning based methods is limited, which means that they cannot explain how an action is performed with detailed information, \ie \emph{who}, \emph{when}, \emph{where} and \emph{how}. Moreover, during their training and testing stages, a video sequence is often assigned a single action category label only. Most of those methods are incapable of detecting concurrent actions in complex videos.

To develop an explainable video understanding framework, the early approaches~\cite{CVIU2001_Intille,ECCV2008_Tran,AAAI2010_Ijsselmuiden,CVPR2011_Morariu,CVPR2011_Brendel} often use first-order logic~\cite{ML2006_Richardson,Book2009_Russell} to predict performed actions or events. Given a video sequence with predefined rules on the spatio-temporal video representation, these algorithms first detect and track concerned objects across the video sequence, and then apply first-order logic to detect occurred actions. Based on the rule-based action definitions, the time and location of the performed action can be detected on the spatio-temporal video representation. In addition, because of the flexibility of logical reasoning framework, concurrent actions can be also detected~\cite{CVPR2011_Morariu}. However, due to low-level image features (\eg motion, location, foreground region) used in early approaches, their explainability and practical applications are still limited. For example, Morariu \etal~\cite{CVPR2011_Morariu} only used the location information (\eg players, ball and loop) to describe the observations of a basketball scene, which is insufficient to describe the ``tumble'' state of a player, because such a semantic state simply cannot be illustrated by locations. 

In this paper, we propose a novel action reasoning framework that uses prior knowledge\footnote{Prior knowledge is the information about a domain that can be used to solve problems in that domain~ \cite{Book2009_Russell}. In this paper, the prior knowledge consists of concerned objects, attributes, relationships and state transition based action definitions.} to explain semantic-level observations of video content changes. In the field of artificial intelligence (AI), an action indicates something done by an agent, and it can be observed by the high-level state transition \cite{Book2009_Russell} from precondition to effect. The precondition defines the states in which the action can be executed, and the effect defines the result of executing the action \cite{Book2009_Russell}. A state here could be an attribute (\eg, a microwave is \emph{closed} or \emph{open}) of an object or a relationship (\eg, a hand is \emph{holding} or \emph{not\_holding} a cup) between two objects. Given the prior knowledge with a set of rules for logical reasoning, performed actions can be detected on the semantic-level video content representation. Consequently, we define two action reasoning models: an \emph{attribute-transition} based and a \emph{relationship-transition} based action reasoning model, or \emph{AAR} and \emph{RAR} model for short, respectively. For example, when the attribute state of a \emph{microwave} is changed from \emph{closed} (precondition) to \emph{open} (effect), it can be inferred as an ``open'' action by the AAR model. Different from low-level features (\eg appearance, location and motion) representation, the semantic-level state is explainable as it can be understood by humans. 

Using AAR and RAR models, we design a spatio-temporal graph for semantic-level video state representation, namely the video graph. Specifically, we extend the scene graph~\cite{IJCV2017_Krishna} in images into a spatio-temporal structure for video graph generation. Each node in our video graph denotes an object and its attributes (including object category, location and its state, \eg \emph{closed} state of a microwave); each edge represents a type of semantic relationship (\eg, a hand is \emph{holding} a cup) between two objects. By detecting and tracking the concerned objects across the video sequence, a spatial-temporal video graph could be generated by sequentially linking the scene graph of each frame throughout the video. Meanwhile, by observing the video graph in temporal order, when the semantic-level video state changes, not only can we recognize the action category with AAR and RAR models, but also explain what happened with detailed state transition information (\ie \emph{who}, \emph{when}, \emph{where} and \emph{why}), as in the example shown in Figure 1. Moreover, since each state transition is independently detected and recognized, multiple concurrent actions can also be handled by the proposed method. To evaluate our method, we re-annotated the CAD dataset with detailed object category, location, attributes, relationships and actions. Experimental results on this dataset demonstrate the effectiveness of the proposed method.

In summary, our contributions are as follows: (1) We propose a novel video action reasoning framework that uses prior knowledge to explain semantic-level video state changes. Compared to previous works, our action reasoning results can be explained by both logical rules and semantic-level observations of video content. (2) We design a video graph representation method for semantic-level video content understanding, and it can detect and provide detailed information for multiple concurrent actions reasoning. (3) We re-annotated the CAD-120 dataset with additional objects, attributes, relationships and actions for empirical studies. Experiments demonstrate the effectiveness of our method in terms of both accuracy and explainability.

\section{Related Work}
\label{sec:relatedwork}

{\bf Action and knowledge representation.}
In the classical planning \cite{Book2009_Russell} problem of AI, an action can be represented by the semantic-level state transition from precondition to effect. For example, in Action Description Language (ADL) \cite{JLC1994_Pednault,Book2009_Russell}, the \emph{load} action can be defined as: 
\begin{equation*}
\begin{split}
Action~( Load (c: Freight, p: Airplane, A: Airport) \\
Precondition: At(c, A) \cap At(p, A) \\
Effect: \neg At(c, A)  \cap In(c, p))
\end{split}
\end{equation*}
where $At(x, A)$ describes whether an object $x$ is at an airport $A$, $In(c, p)$ denotes whether a freight $c$ is in an airplane.
Given the knowledge of a certain domain, such action representation lifts the level of reasoning from propositional logic
to a restricted subset of first-order logic \cite{Book2009_Russell}. Besides, for the logical and flexible knowledge representation,
semantic network \cite{Book2014_Sowa,Book2009_Russell,Book2010_Poole} is capable of representing objects, attributes of object and relations among relevant objects in real world. Here, the semantic network is a graph based knowledge representation method in AI, where a node represents an object and an edge describes the relationship between two different objects. Inspired by the semantic network for knowledge representation and logical reasoning,
we design a video graph for detailed semantic-level video content representation by extending the scene graph~\cite{IJCV2017_Krishna} into a spatio-temporal structure. By observing the state changes over time, performed actions can be detected and explained by logical rules.

{\bf Video action recognition.}
Without rules for logical reasoning, many approaches often employ hand-crafted \cite{BMVC2008_Klaser,CVPR2008_Cordelia,ICCV2013_Wang,TPAMI2017_Liu} or deep-learned features \cite{NIPS2014_Simonyan,TPAMI2018_Wang,CVPR2017_Feichtenhofer,ICCV2017_Lee,CVPR2018_Wang,CVPR2018_Feichtenhofer} of appearance and motion for action recognition. Recently, researchers attempt to use the semantic-level state changes~\cite{CVPR2013_Fathi,TIST_Fire2015,CVPR2016_Wang,ICCV2017_Liu,ICCV2017_Alayrac,ECCV2018_Wang} for video analysis. For example, Liu \etal \cite{ICCV2017_Liu} adopted unary fluents to represent attributes of a single object, and binary fluents for two objects in egocentric videos, and then they used LSTM~\cite{SSL2012_Graves} to recognize which action is performed. In addition, Recurrent Neural Networks (RNN)~\cite{CVPR2016_Jain} or Graph Convolutional Networks (GCN)~\cite{ICLR2017_Kipf,ECCV2018_Wang,ECCV2018_Guo,ECCV2018_Qi} is used for structured video representation and action recognition in 2D or 3D scenes. Due to the absence of rules for logical reasoning, the explainability of these methods is limited.

Early logical reasoning based methods~\cite{CVIU2001_Intille,ECCV2008_Tran,AAAI2010_Ijsselmuiden,CVPR2011_Morariu,CVPR2011_Brendel} often use logical rules to explain the low-level features, such as motion, location and trajectories. Tran \etal~\cite{ECCV2008_Tran} used Markov logic network~\cite{ML2006_Richardson} and first-order logic for event recognition in surveillance domains. By observing the location of cars and pedestrians in the scenes, a set of actions can be inferred, such as ``door opening'' and ``car leaving''. Brendel \etal~\cite{CVPR2011_Brendel} introduced a probabilistic event logic method for interval-based event recognition, and they used a manually defined knowledge base to interpret low-level histogram of gradients (HoG) and the histogram of flow (HoF) features. Different from previous works, our method uses logical rules to explain semantic-level observations of video content changes.

{\bf Scene graph representation in images.}
To understand and describe the visual world in a single image,
Krishna \etal \cite{IJCV2017_Krishna,CVPR2018_Krishna} proposed a scene graph representation for rich image content cognition.
In \cite{IJCV2017_Krishna,CVPR2018_Krishna}, an image is represented by a set of objects, attributes, and relationships.
Given an input image, the object is represented by the its category, location, and attributes that denote detailed object information (\eg \emph{age} and \emph{gender} of a man).
Relationship connecting two different objects represents the semantic relation between the subject and object (\eg, \emph{a man is holding a cup}).
To further exploit robust relationship prediction with limited training samples,
Lu \etal \cite{ECCV2016_Lu} proposed a visual relationship model with language priors.
Dai \etal \cite{CVPR2017_Dai} developed a Deep Relational Network (DR-Net),
which integrates a variety of cues: object categories, appearance, spatial configurations, and the statistical relations. 

Instead of using static image scene description, 
we track the concerned objects with their attributes and relationships by going frame-wise through the whole video sequence to construct a video graph,
which is used to represent the video state (attribute and relationship) transitions.

\section{Explainable Action Analysis}
\label{sec:proposed}
In this section, we introduce the proposed problem definition, video graph generation, and how to reason about performed actions by integrating prior knowledge with logical rules.


\begin{figure*}[t]
	\centering
	\includegraphics[width=0.9\textwidth,height=0.22\textheight]{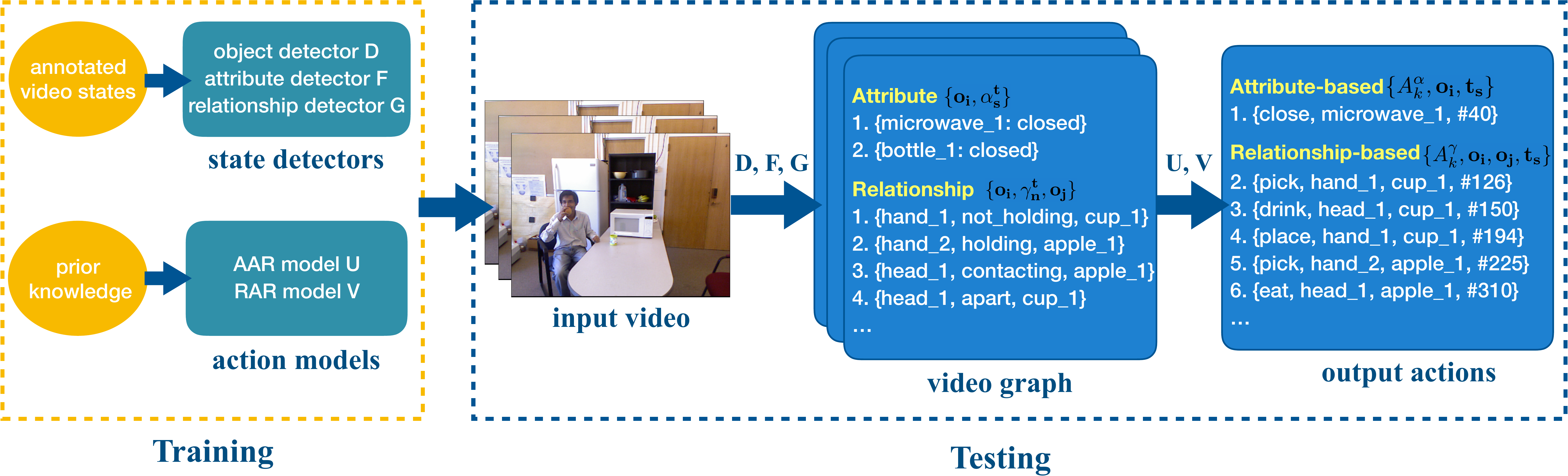}
	\caption{An overview of the proposed video action reasoning framework. AAR denotes attribute-based action reasoning model and RAR represents relationship-based action reasoning model, $t_s$ is the state transition time. Based on the prior knowledge and video graph representation, performed actions can be detected and explained with detailed information.}
	\label{fig_framework}
\end{figure*}
\subsection{Problem definition}
\label{subsec_difinition}
Given a video of the target domain (such as daily life),
we define the {\bf prior knowledge} in this domain as a set of concerned object categories $\Mat{o} = \{o_1, \cdots, o_M\}$,
associated attributes $\Mat{\alpha} = \{\alpha_1, \cdots, \alpha_S\}$,
potential relationships $\Mat{\gamma} = \{\gamma_1, \cdots, \gamma_N\}$ between each pair of objects,
and possible actions $\Mat{A} = \{A_1, \cdots, A_K\}$ that might be performed among these objects.
According to the state transition \cite{JLC1994_Pednault,Book2009_Russell,Book2010_Poole} based action representation,
two types of video state are considered for video action reasoning: \emph{attribute-based state} and \emph{relationship-based state}.
Given a target video domain, the number of concerned objects and states is limited,
and thus we can build a complete knowledge base\footnote{A knowledge base represents facts about the world that are stored by an agent~ \cite{Book2009_Russell}.} \cite{Book2009_Russell} to explain all concerned state transitions.

{\bf Attribute.}
Let $W_{o_i}^{\alpha_s}$ be the state of object $o_i$ with attribute $\alpha_s$,
where $i \in \{1, \cdots, M\}$ denotes the $i$-th concerned object category,
$s \in \{1, \cdots, S\}$ denotes the $s$-th concerned attribute.
Then an attribute-based action $A^\alpha$ can be defined as a valid attribute transition $W_{o_i}^{\alpha_p} \rightarrow W_{o_i}^{\alpha_e}$,
where $\alpha_p$ is the initial attribute as precondition and $\alpha_e$ is the effect of executing the action.

{\bf Relationship.}
Similarly, let $W_{o_i, o_j}^{\gamma_n}$ be the state of object $o_i$ and $o_j$ with semantic relationship $\gamma_n$.
$n \in \{1, \cdots, N\}$ denotes the $n$-th concerned relationship.
The relationship-based action $A^\gamma$ can be defined as $W_{o_i,o_j}^{\gamma_p} \rightarrow W_{o_i,o_j}^{\gamma_e}$ ($i \neq j$),
where $W_{o_i,o_j}^{\gamma_p}$ and $W_{o_i,o_j}^{\gamma_e}$ denote the precondition and effect of a relationship, respectively.

{\bf Video graph.}
Given a video $\Mat{I} = \{I^1, \cdots, I^T\}$ with $T$ frames,
we assume there are $m$ concerned objects $\{\Mat{o}_1, \cdots, \Mat{o}_m\}$ in this video domain.
Each object $\Mat{o}_i$ in frame $I^t$ is represented by its category $o_{\tilde{i}}$ and a bounding box location $\Mat{b}^t$
as $\Mat{o}_i^t =\{o_{\tilde{i}}, \Mat{b}_i^t\}$.
To describe the video state transitions, a video graph $\Mat{\mathcal{G}} = \{\mathcal{G}^1, \cdots, \mathcal{G}^T\}$ is defined to represent the attribute and relationship on each object, where $\mathcal{G}^t \in \Mat{\mathcal{G}}$ is a scene graph, which denotes the state of video frame $I^t$.

{\bf Attribute-based action definition.}
Let $W_{o_{\tilde{i}}}^{\alpha_s^t}$ denote the attribute state of object $\Mat{o}_i$ at frame $I^t$,
$W_{o_{\tilde{i}}, o_{\tilde{j}}}^{\gamma_s^t}$ represent the relationship state between object $\Mat{o}_i$ and $\Mat{o}_j$ at frame $I^t$.
When $W_{o_{\tilde{i}}}^{\alpha_s^t} \neq W_{o_{\tilde{i}}}^{\alpha_s^{t+1}}$,
an attribute-transition has occurred, namely $W_{o_{\tilde{i}}}^{\alpha_s^t} \rightarrow W_{o_{\tilde{i}}}^{\alpha_s^{t+1}}$,
and the state transition time (\ie when) is marked as $t+1$.
Then the performed action $A^\alpha$ can be explained by the attribute-transition based action definitions,
and the location (\ie where) of performed action is represented by $\Mat{b}_i^{t+1}$ of object $\Mat{o}_i$.

{\bf Relationship-based action definition.}
Similarly, the relationship-based action can be defined by relationship-transition
$W_{o_{\tilde{i}}, o_{\tilde{j}}}^{\gamma^t} \rightarrow W_{o_{\tilde{i}}, o_{\tilde{j}}}^{\gamma^{t+1}}$,
and the locations can be denoted by $\Mat{b}_i^{t+1}$ and $\Mat{b}_j^{t+1}$ of object $\Mat{o}_i$ and $\Mat{o}_j$, respectively.

{\bf Video action reasoning.}
Given a video $\Mat{I}$ and the prior knowledge about concerned object categories $\Mat{o}$, attributes $\Mat{\alpha}$, relationships $\Mat{\gamma}$,
and action definitions $\Mat{A}$ with a set of state transition
$W_{o_i}^{\alpha_p} \rightarrow W_{o_i}^{\alpha_e}$ and $W_{o_i,o_j}^{\gamma_p} \rightarrow W_{o_i,o_j}^{\gamma_e}$ ($i \neq j$),
based on the semantic-level video graph representation, our target is to recognize and reason a set of performed actions $A^\alpha$ and $A^\gamma$ when $W_{o_{\tilde{i}}}^{\alpha_s^t} \neq W_{o_{\tilde{i}}}^{\alpha_s^{t+1}}$ or $W_{o_{\tilde{i}}, o_{\tilde{j}}}^{\gamma^t} \neq W_{o_{\tilde{i}}, o_{\tilde{j}}}^{\gamma^{t+1}}$, respectively.

\subsection{Overview of the Proposed Method}
Similar to previous works~\cite{CVPR2011_Morariu,CVPR2011_Brendel} for explainable video analysis, we use manually defined prior knowledge as the logical rules for action reasoning. In addition, to overcome the limitation of low-level features used in ~\cite{CVPR2011_Morariu,CVPR2011_Brendel}, we design a semantic-level video content representation method, namely video graph. \fig \ref{fig_framework} shows the overview of the proposed action reasoning framework.

{\bf Training stage.}
The training stage includes two aspects: (1) the state detectors consist of the object detector $D$, attribute detector $F$ (\eqn \ref{eq_attribute}) and relationship detector $G$ (\eqn \ref{eq_relation}).
Similar to the scene graph \cite{IJCV2017_Krishna,ECCV2016_Lu,CVPR2017_Dai} generation in still images, those state detectors are trained on annotated video states; (2) The action models include the Attribute-based Action Reasoning (AAR) model $U$ (\eqn \ref{eq_act-attr}) and Relationship-based Action Reasoning (RAR) model $V$ (\eqn \ref{eq_act-rel}). Different from many existing methods~\cite{CVPR2017_Carreira,CVPR2018_Wang,CVPR2018_Feichtenhofer} that learn the action model on well-annotated video clips (a single action label for a video clip), the state transition based action models are learned on the given prior knowledge (semantic-level action definitions).

{\bf Testing stage.}
Given a video sequence, we first employ the trained object detector $D$, attribute detector $F$, and relationship detector $G$ to
detect the concerned objects, attributes and relationships on each frame. By tracking those objects across different video frames,
a video graph is generated for semantic-level state representation.
Then the AAR model $U$ and RAR model $V$ are used to reason about the performed actions with attribute or relationship transitions, respectively.	Since each state transition is explained by the logical rules independently, our method is able to obtain detailed action information from these state transitions, and it can be also used to detect multiple actions in complex videos, including concurrent actions.

\subsection{Video Graph Generation}
\label{sec_vide-graph}
By assuming the objects and their locations are available \cite{CVPR2016_He,AAAI2017_Szegedy}, as in scene graph \cite{IJCV2017_Krishna,CVPR2017_Xu,CVPR2017_Dai}, attribute and relationship detectors can be learned on still video frames. 


{\bf Attribute detector.}
Let $\Mat{o} = \{\Mat{o}_1, \cdots, \Mat{o}_m\}$ denote the $m$ concerned objects in the whole video,
$\Mat{o}_i^t =\{o_{\tilde{i}}, \Mat{b}_i^t\}$ be the $i$-th object $\Mat{o}_i^t$ with category $o_{\tilde{i}}$ and location $\Mat{b}_i^t$ in frame $I^t$.
The same attribute category may show very different visual appearances to different objects. For example, the \emph{open} states of a microwave and a bottle are quite different. In order to ensure the semantic representativeness capability of the trained attribute classifier, the object label needs to be incorporated into the attribute model. Given a set of images with annotated objects and attributes, the object attribute $\alpha_s^t$ with the detector $F$ can be learned as:
\begin{equation}
\label{eq_attribute}
\begin{aligned}
F(\alpha_s^t, \Mat{\Delta} | <\Mat{o}_i^t, I^t>) = \Mat{z}_s^\alpha [\phi(\Mat{b}_i^t, I^t) \delta(w2v(o_{\tilde{i}}), \Mat{\omega}^\alpha)] + d_s^\alpha
\end{aligned}
\end{equation}
where $\Mat{\Delta}=\{\Mat{z}_s^\alpha, d_s^\alpha\}$ denotes the learned parameters to predict attribute probabilities,
$s \in \{1, \cdots, S\}$ represents $s$-th concerned attribute in the target domain.
$\phi(\Mat{b}_i^t, I^t)$ is the extracted visual feature via convolutional neural networks in location $\Mat{b}_i^t$ of the frame $I^t$.
$w2v(o_{\tilde{i}})$ is the one-hot feature vector of object category with dimension $M$.
$\Mat{\omega}^\alpha$ is the learned parameter for feature selection, and
$\delta(w2v(o_{\tilde{i}}), \Mat{\omega}^\alpha)$ is the learned weights from object category.

{\bf Relationship detector.}
Different from detailed image understanding in scene graph \cite{IJCV2017_Krishna}, we only focus on the relationships that can be used to generate concerned actions.
Similar to attribute detector, we also consider the object categories as in \eqn \ref{eq_attribute}.
Let $\Mat{o}_i^t =\{o_{\tilde{i}}, \Mat{b}_i^t\}$ denote the subject and $\Mat{o}_j^t =\{o_{\tilde{j}}, \Mat{b}_j^t\}$ represent the object in frame $I^t$.
The spatial relationship $\gamma_n^t$ can be calculated by the detector $G$ \cite{ECCV2016_Lu} as:
\begin{equation}
\label{eq_relation}
\begin{aligned}
G(\gamma_n^t, \Mat{\Theta} | <\Mat{o}_i^t, \Mat{o}_j^t, I^t>)  \\
= \Mat{z}_n^\gamma [\varphi(\Mat{b}_i^t, \Mat{b}_j^t, I^t) \delta([w2v(o_{\tilde{i}}), w2v(o_{\tilde{j}})], \Mat{\omega}^r)]+d_n^\gamma
\end{aligned}
\end{equation}
where $\Mat{\Theta}=\{\Mat{z}_n^\gamma, d_n^\gamma\}$ denotes the learned parameters to predict relationship probabilities,
$n \in \{1, \cdots, N\}$ represents $\text{n-th}$ relationship in the target domain.
$\varphi(\Mat{b}_i^t, \Mat{b}_j^t, I^t)$ is the feature extracted from the union region of the $\Mat{b}_i^t$ and $\Mat{b}_j^t$ boxes in frame $I^t$.
$w2v(o_{\tilde{i}})$ and $w2v(o_{\tilde{j}})$ denote the one-hot vectors of subject category $\Mat{o}_{\tilde{i}}$ and object category $\Mat{o}_{\tilde{j}}$, respectively.
$[w2v(o_{\tilde{i}}), w2v(o_{\tilde{j}})]$ represents the concatenated object category vector.
$\Mat{\omega}^\gamma$ is the learned parameter for feature selection.
$\delta([w2v(o_{\tilde{i}}), w2v(o_{\tilde{j}})], \Mat{\omega}^\gamma)$ is the extracted weights from both subject and object categories.

{\bf Video graph generation and refinement.}
Based on detected objects, attributes and relationships, semantic-level state in each individual frame can be represented by the generated scene graph.
By tracking each object across all the video frames, a video graph is further generated.
Then state transitions can be detected by observing the attribute and relationship changes over time,
as illustrated in \fig \ref{fig_video_graph}. In addition, since the proposed video graph is a general and flexible video representation method, it can be also used in other computer vision tasks, such as video summarization~\cite{TMM2012_Event,TMM2012_Movie} and video captioning~\cite{TCSVT2018_Xu}.

In a complex video scene,
the predicted attributes and relationships are sometimes inaccurate, which need to be refined for more robust video graph generation.
With the assumption that state changes in a video sequence should be consistent, a sliding window with a width $\theta$ is utilized to improve the quality of the generated video graph.
When the same state is not continuously detected, the detection is considered inaccurate, and then the latest accurate value is assigned to the inaccurate state. 
Figure 3 shows an example of refinement using a width of 3, where 0 denotes one state and 1 represents another possible state.

\begin{figure}[t]
	\centering
	\includegraphics[width=0.4\textwidth]{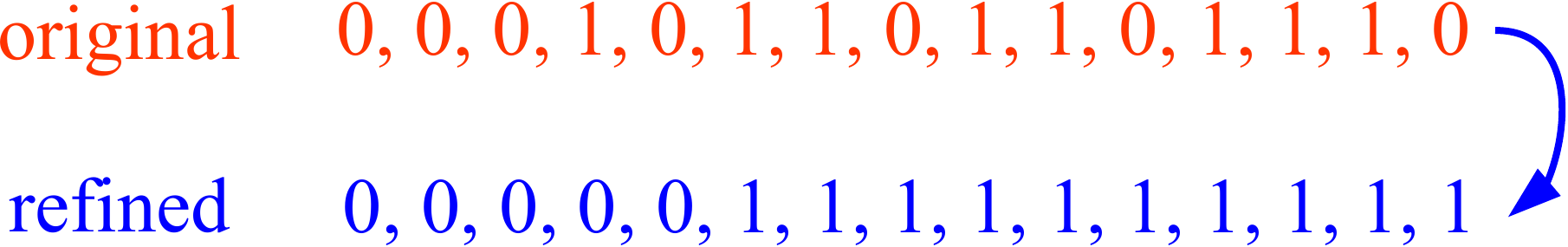}
	\caption{An example of state refinement (width $\theta=3$).}
	\label{fig_state_refined}
\end{figure}

\subsection{Explainable Action Reasoning Models}
Suppose that the environment state will not change unless an action is performed. In the real world, a complex action can contain different objects, attribute and relationship transitions. For those actions, they can be divided into a set of atomic propositions with first-order logic \cite{Book2009_Russell,Book2014_Sowa}. For example, the ``having\_meal'' activity can be defined by two atomic actions as: $\emph{eat} \wedge \emph{drink}$. Therefore, in order to clearly introduce the proposed method, we mainly focus on the atomic action reasoning model that involves only one attribute or relationship transition in this paper.

{\bf Attribute-based action reasoning (AAR) model.}
The AAR model is used to detect the attribute-transition of a node in video graph,
such as ``open a microwave'' (the attribute changes from \emph{closed} to \emph{open}).
Considering both object category and state transition, the attribute-based action $A_k^\alpha$ is formulated by a projection function $U$ as:
\begin{equation}
\begin{aligned}
U(A_k^\alpha, \Mat{\Phi} | <o_i, \alpha_p, \alpha_e>) \\
=\Mat{w}_k^\alpha [w2v(o_i), w2v(\alpha_p), w2v(\alpha_e)]+b_k^\alpha
\end{aligned}
\label{eq_act-attr}
\end{equation}
where $\Mat{\Phi} = \{\Mat{w}_k^\alpha, b_k^\alpha\}$ is the learned parameter,
$k \in \{1, \cdots, K\}$ represents $k$-th concerned action in the target video domain.
$\alpha_p$ is a attribute of precondition, and $\alpha_e$ is the attribute of effect.

{\bf Relationship-based action reasoning (RAR) model.}
In contrast, the RAR model is used to detect the relation-transition of an edge in video graph,
such as ``hand picks a cup'' (spatial relationship between \emph{hand} and \emph{cup} changes from \emph{not\_holding} to \emph{holding}).
Similar to the attribute-based action model, we also build a conjunction of the subject and object categories with relationship-transition to distinguish the action on different object categories.
Then relationship-based action $A_k^\gamma$ can be formulated by a projection function $V$ as:
\begin{equation}
\begin{aligned}
\small
V(A_k^\gamma, \Mat{\Psi} | <o_i, o_j, \gamma_p, \gamma_e>)  \\
= \Mat{w}_k^\gamma [w2v(o_i), w2v(o_j), w2v(\gamma_p), w2v(\gamma_e)]+b_k^\gamma
\end{aligned}
\label{eq_act-rel}
\end{equation}
where $\Mat{\Psi} = \{\Mat{w}_k^\gamma, b_k^\gamma\}$ is the learned parameter, and $k \in \{1, \cdots, K\}$ denotes $k$-th concerned action in the target video domain.
$\gamma_p$ is a relationship of precondition, while $\gamma_e$ is the relationship of effect.

{\bf Learning action models from prior knowledge.}
Since (1) the number of concerned objects and state transitions in a target video domain is often limited \cite{JLC1994_Pednault} and
(2) all of the potential actions can be represented by different state transitions, an appropriate action model can be learned from a complete knowledge base (\ie state-transition based action definitions)  \cite{Book2009_Russell,Book2010_Poole}.

\section{Experiments}
In this section, we first introduce the dataset and prior knowledge definition in this work, and then describe the implementation details of our method. In the next, we report the accuracy of the video graph generation and action reasoning results, and demonstrate the explainability of our method. Finally, we discuss the potential extensions of the proposed method.

\subsection{Dataset and Prior Knowledge Definition}
For explainable video action reasoning, the annotated objects, attributes, relationship and actions are employed to validate the effectiveness of our method. However, available datasets do not satisfy these requirements. Therefore, we construct a new dataset by re-annotating the CAD-120 dataset \cite{IJRR_Hema} with detailed object locations, attributes, relationships and actions.  
The original CAD-120 is a RGBD dataset (we only use the RGB images for action analysis, depth information is ignored in this work) captured by Microsoft Kinect sensor \cite{CVPR2011_Shotton}, which focuses on the human activity of daily life. Each image of CAD-120 has resolution of $640 \times 480$ and the whole dataset contains 124 video sequences of 10 different high-level activities (such as arranging objects, having meal, taking food) performed by 4 different subjects, and each activity was performed 3 or 4 times.
In addition, each action is carried out on different objects, such as ``pick a cup'' and ``pick a box''. 

\begin{table}[!t]
	\small
	\centering
	\caption{Attributes of an object.}
	\label{tbl_attr}
	\begin{tabular}{c|c}
		\toprule
		Attribute & Object                                     \\ \hline
		closed/open    & medicine-box, microwave, bottle    \\ 
		\bottomrule
	\end{tabular}
\end{table}
\begin{table}[!t]
\centering
\caption{Attribute-based action definitions.}
\label{tbl_act_attr}
	\resizebox{1\columnwidth}{!}{
	\begin{tabular}{c|c|c}
	\toprule
	Action & Attribute Transition         & Object                          \\ \hline
	close  & open $\rightarrow$ closed   & medicine-box, microwave, bottle \\ \hline
	open   & closed $\rightarrow$ open & medicine-box, microwave, bottle \\ 
	\bottomrule
	\end{tabular}
	}
\end{table}

As an extension, the re-annotated CAD-120 dataset \cite{IJRR_Hema} consists of 551 video clips with 32327 frames. For the purpose of extending, 
10 potential actions (include \emph{null} action that means nothing changed/happened) have been defined in the target domain (daily life), as well as a set of concerned objects, attributes and relationships.
Since we often describe an action (as well as attribute and relationship) on different objects with the same term in our daily life, such as ``\emph{pick} a book'' and ``\emph{pick} an apple'', we also follow this tradition in the proposed framework.

\begin{table}[!t]
	\centering
	\caption{Relationships between two objects.}
	\label{tbl_rel}
	\resizebox{0.95\columnwidth}{!}{
	\begin{tabular}{c|c|c}
		\toprule
		Relationship           & Subject            & Object                                                                  \\ \hline
		holding / not\_holding & hand               & \makecell{box, medicine-box, bowl, \\
																cup, book, cloth, remote, \\
																apple, bottle, plate}                                         \\   \hline
		contacting / apart     & head               & bottle, bowl, cup, apple                                                \\ \hline
		containing / separate  & microwave          & bowl, cup, cloth, box                                                   \\ 
		\bottomrule
	\end{tabular}
	}
\end{table}
\begin{table}[!t]
\small
\centering
\caption{Relationship-based action definitions. }
\label{tbl_act_rel}
\resizebox{1\columnwidth}{!}{
\begin{tabular}{c|c|c|c}
\toprule
Action          & Relationship Transition            & Subject            & Object                                                    \\ \hline
pick            & not\_holding $\rightarrow$ holding & hand               & \makecell{box, medicine-box, \\
																			bowl, cup, book, \\
																			cloth, remote, \\
																			apple, bottle, plate} \\ \hline
place           & holding $\rightarrow$ not\_holding & hand               & \makecell{box, medicine-box, \\
																			bowl, cup, book, \\
																			cloth, remote, \\
																			apple, bottle, plate}                                     \\ \hline
drink           & apart $\rightarrow$ contacting     & head               & cup, bottle                                               \\ \hline
eat             & apart $\rightarrow$ contacting     & head               & apple, bowl                                               \\ \hline
micr\_food      & separate $\rightarrow$ containing  & microwave          & cup, box, bowl                                            \\ \hline
take\_food      & containing $\rightarrow$ separate  & microwave          & cup, box, bowl                                            \\ \hline
clean           & separate $\rightarrow$ containing  & microwave          & cloth                                                     \\ 
\bottomrule
\end{tabular}
}
\end{table}

\begin{table*}[!ht]
	\centering
    \caption {State recognition accuracy of the proposed method with (w) or without (w/o) object categories.}
	\resizebox{0.8\textwidth}{!}{
    \begin{tabular}{c|c|c|c|c|c|c|c|c|c}
    \toprule
    \bf{State}    & \bf{closed} & \bf{open} & \bf{holding} & \bf{not\_holding} & \bf{contacting} & \bf{apart}  & \bf{containing} 
                                                                                                   & \bf{separate} & \bf{Overall}  \\ \hline
    w/o obj   & \bf{0.99}   & 0.70    & \bf{0.86}    & 0.91         & 0.59       & 0.60   & 0.93       & 0.67     & 0.82     \\ \hline
    w obj & 0.98   & \bf{0.74}    & 0.82    & \bf{0.96}    & \bf{0.80}  & \bf{0.96}  & \bf{0.95}   & \bf{0.96}   & \bf{0.94}     \\ \bottomrule
    \end{tabular}
	}
	\label{tbl_state_result}
\end{table*}
\begin{figure}[t]
	\centering
	\subfigure[False attribute as \emph{open}.]{
		\label{fig_fail_attr}
		\includegraphics[width=0.20\textwidth]{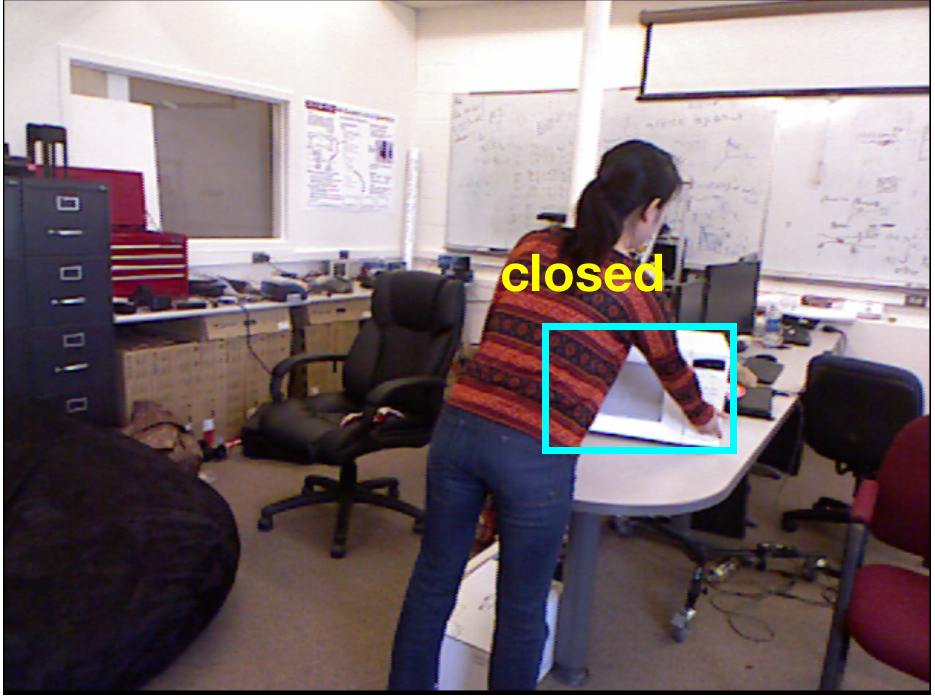}}
	\subfigure[False relationship as \emph{holding}. ]{
		\label{fig_fail_rel}
		\includegraphics[width=0.20\textwidth]{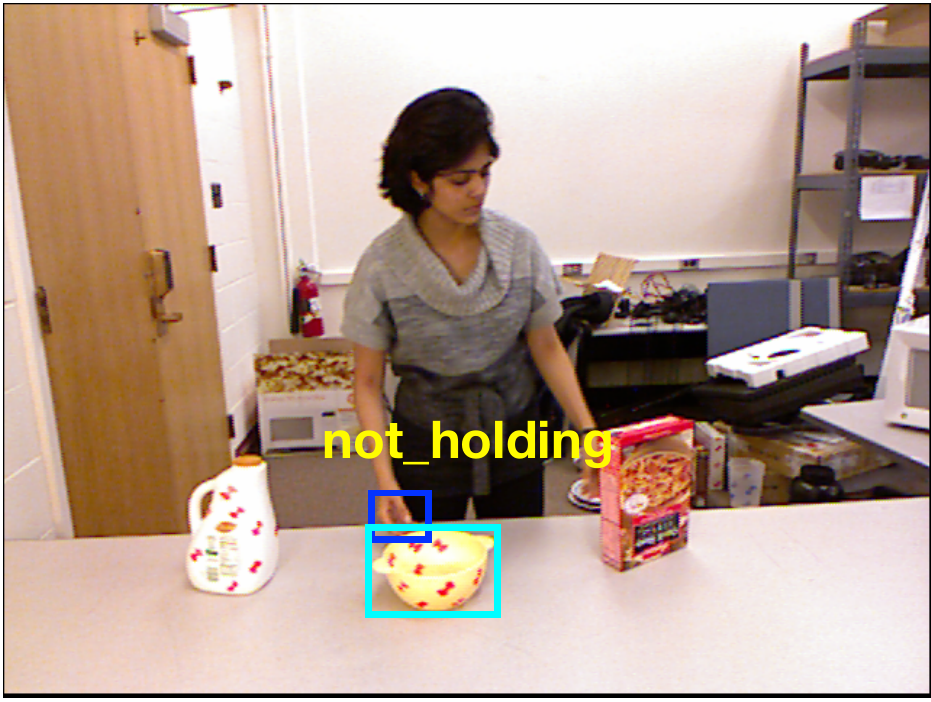}}
	\caption{Failure cases.}
	\label{fig_failure}
\end{figure}

\tab 1-4 represent the prior knowledge about the target domain used in this work and
the action models can be trained from \tab \ref{tbl_act_attr} and \tab \ref{tbl_act_rel} without any annotated videos.
More explicitly, \tab \ref{tbl_attr} and \tab \ref{tbl_rel} are the concerned object categories, attributes and relationships;
\tab \ref{tbl_act_attr} and \tab \ref{tbl_act_rel} are the defined actions based on valid attribute-transition and relationship-transition, respectively.
When the state does not change or the state transitions are not contained in \tab \ref{tbl_act_attr} and \tab \ref{tbl_act_rel}, \emph{null} is used to represent such conditions as denoting nothing happened. In summary, there are 13 object categories, 2 attributes, 6 relationships, 12 attribute-based transitions, 72 relationship-based transitions, and 10 actions in total. Based on the prior knowledge on this video domain, we can generate video graphs for semantic-level video content understanding and further action reasoning.

\subsection{Implementation Details}
Based on the manually labeled data,
the attribute and relationship detectors are trained with a VGG-16 model \cite{CoRR2014_Karen}, respectively.
The learning rate is set to $10^{-5}$ for both attribute and relationship detectors.
For the training of AAR and RAR action reasoning models, the learning rate is set to 0.01.
In addition, for all detectors, the categorical cross-entropy is used as the loss function and
Adam optimizer \cite{CoRR2014_Adam} is used for optimization.
Besides, the smoothing width $\theta$ is empirically set to 5 for robust video graph generation.


\begin{figure*}[!t]
	\centering
	\includegraphics[width=0.8\textwidth, height=0.22\textheight]{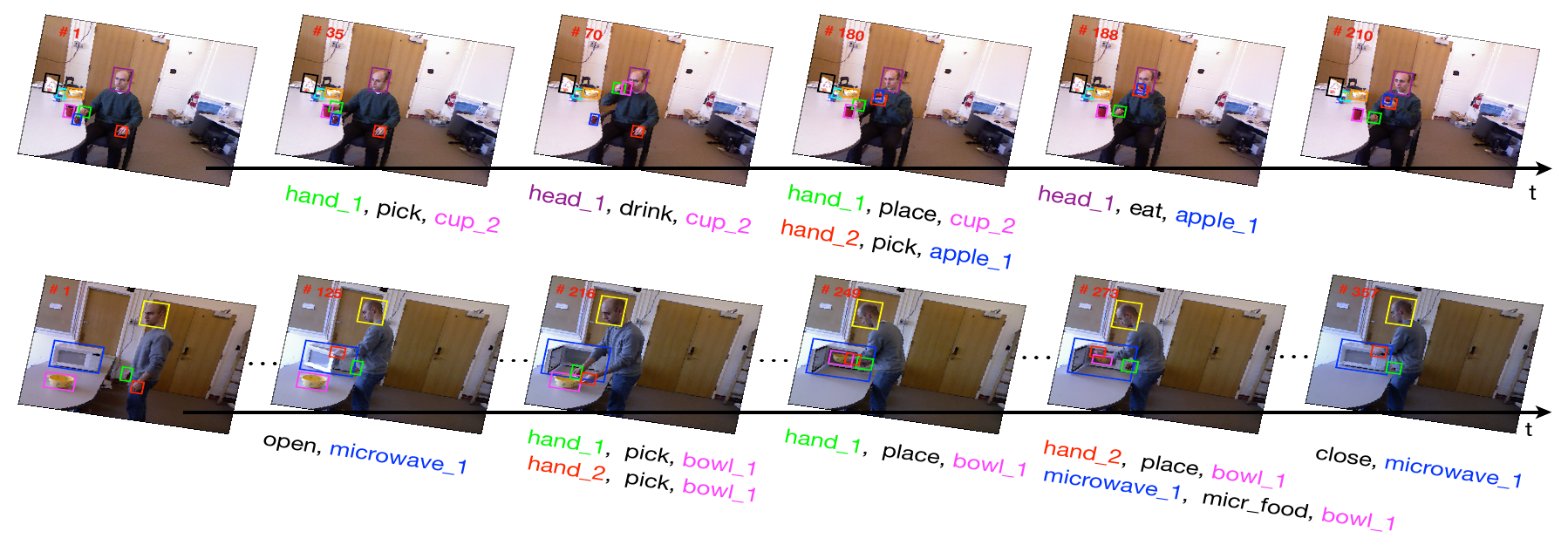}
	\caption{Examples of video action reasoning with detailed state transition information in complex videos.}
	\label{fig_act_video}
\end{figure*}
\subsection{Video Graph Generation}
Given a video sequence with manually annotated objects on a few frames, we use a tracking algorithm~\cite{CVPR2017_Lukezic} to estimate the locations of each object in the rest of video. By using the described methods (as in Sec. \ref{sec_vide-graph}) for scene graph generation on each frame, a video graph is further generated. In the state transition based action reasoning framework, when an action occurs, it can be detected by the temporal attribute or relationship changes (\ie state transitions) in the generated video graph. Because our action models are learned from the prior knowledge, the performance of our method depends on the accuracy of state detection. If the more accurate video state detection results are provided, more accurate action reasoning results can be obtained. 

To demonstrate the accuracy of the state detection, we report both the attribute and relationship recognition accuracy. In many cases, because the same attribute and relationship often exist in different object categories, it can be very diverse, such as ``a hand is holding a box'' and ``a hand is holding a cup''. With the consideration of the object categories in attribute and relationship modeling, the overall performance has been effectively improved, as the results shown in \tab \ref{tbl_state_result}. 

For some challenging scenes, such as heavy occlusion and inconspicuous conditions, it is difficult to accurately predict the states. As the example shown in \fig \ref{fig_fail_attr}, when the \emph{microwave} is occluded by the girl, its attribute is incorrectly predicted as \emph{open}. For another example shown in \fig \ref{fig_fail_rel}, the ground truth relationship is \emph{not\_holding},
but it is improperly detected as \emph{holding} since the hand is very close to the bowl. In practice, although the accuracy of state detection is not perfect, it still provides important information for action reasoning, see Sec. \ref{subsec_action_reason} and Sec. \ref{subsec_action_acc}. 

\subsection{Explainable Action Reasoning}
\label{subsec_action_reason}


By observing each state transition in the video graph over time, performed actions can be detected and explained by the rule-based action reasoning models (\ie AAR and RAR). Moreover, since each state transition in the video graph is detected and explained independently, detailed state transition information (\ie \emph{who}, \emph{when}, \emph{where} and \emph{how}) can be obtained from the video graph, and multiple concurrent actions can also be detected.

To illustrate the advantages of the proposed method, action reasoning results on two long video sequences are reported.
As shown in \fig \ref{fig_act_video},
it can be seen that all of the concerned objects are represented in these two videos.
When their attributes or relationships change, performed actions are detected and marked with relative objects at the video frames where the state transitions occur. In the first row of \fig \ref{fig_act_video},
the man ``picks cup\_2'' (\emph{not\_holding} to \emph{holding} between \emph{hand\_1} and \emph{cup\_2}) with ``hand\_1'' at frame 35;
then he uses ``cup\_2'' to ``drink'' (\emph{apart} to \emph{contacting} between \emph{head\_1} and \emph{cup\_2}) at frame 70.
Later, at frame 180,
the man ``places cup\_2'' (\emph{holding} to \emph{not\_holding} between \emph{hand\_1} and \emph{cup\_2}) and ``picks apple\_1'' with ``hand\_2''
(\emph{not\_holding} to \emph{holding} between \emph{hand\_2} and \emph{apple\_1}) at the same time.
Finally, he ``eats apple\_1'' (\emph{apart} to \emph{contacting} between \emph{head\_1} and \emph{apple\_1}) at frame 188.
Note that although some objects are irrelevant to any actions, such as the \emph{bowl} (marked in yellow rectangle) and \emph{bottle} (marked in black rectangle),
they are still need to be considered in the action reasoning stage. This is because we do not know what will happen at the beginning of an input video, we need to monitor all the objects.

Similarly, the second row of \fig \ref{fig_act_video} shows another action reasoning result.
The man ``opened microwave\_1'' (\emph{closed} to \emph{open} of \emph{microwave\_1}) at frame 125,
then the concurrent actions ``hand\_1 picks bowl\_1'' and ``hand\_2 picks bowl\_1'' are performed at frame $216$,
another concurrent actions ``hand\_2 places bowl\_1'' and ``microwave\_1 micro\_food bowl\_1'' are performed at frame $273$.
In the end, the man ``closes microwave\_1'' at frame 367.

As we can see, our method can provide a video analysis strategy as in the logical manner of thinking by humans. Different from previous logical reasoning methods ~\cite{CVIU2001_Intille,ECCV2008_Tran,AAAI2010_Ijsselmuiden,CVPR2011_Morariu,CVPR2011_Brendel} and semantic-level video action recognition algorithms~\cite{CVPR2013_Fathi,TIST_Fire2015,CVPR2016_Wang,ICCV2017_Liu,ICCV2017_Alayrac,ECCV2018_Wang}, the explainability of our method is supported by both logical rules and semantic-level video content understanding. Besides, \fig \ref{fig_act_video} also demonstrates that the proposed method can detect multiple concurrent actions in complex videos, and it can also provide detailed action information to explain how those actions are executed.

\begin{table*}
	\centering
	\caption {Comparison between the proposed method and TSN \cite{TPAMI2018_Wang} for single action recognition.}
	\resizebox{0.80\textwidth}{!}{
	\begin{tabular}{c|c|c|c|c|c|c|c|c|c|c|c}
		\toprule
      \bf{Action}        & \bf{null} & \bf{open} & \bf{close} & \bf{pick} & \bf{place} & \bf{drink} & \bf{eat}  
									      & \bf{micr\_food} & \bf{take\_food} & \bf{clean} & \bf{Overall} \\  \hline
    \bf{Video number}    & 161       & 48        & 40    & 109  & 100  & 31    & 28   & 10   & 12    & 10  & 551     \\ \hline
    TSN\_RGB        & 0.92      & 0.46      & 0.45      & 0.53      & 0.57       & 0.74      & 0.21      & 0         & 0         & 0.33      & 0.42  \\ \hline
    TSN\_Flow      & 0.91      & 0.69      & 0.70      & 0.81      & \bf{0.92}  & 0.84      & 0.39      & 0.80      & 0.33      & 0.67      & 0.71  \\ \hline
    TSN\_Fusion     & \bf{0.96} & 0.77      & 0.50      & \bf{0.88} & 0.87       & 0.86      & 0.20      & \bf{1.00} & \bf{0.67} & \bf{1.00} & 0.77  \\ \hline
    Ours             & 0.84      & \bf{0.83} & \bf{0.88} & 0.68      & 0.86       & \bf{1.00} & \bf{0.89} & 0.50      & \bf{0.67} & 0.83      & \bf{0.80} \\ \bottomrule
	\end{tabular}
    }
    \label{tbl_tsn_our}
\end{table*}

\subsection{Action Recognition Accuracy}
\label{subsec_action_acc}

Note that there are essential differences between the proposed action reasoning approach and many deep learning based action recognition methods \cite{NIPS2014_Simonyan,TPAMI2018_Wang,CVPR2017_Feichtenhofer,ICCV2017_Lee,CVPR2018_Wang,CVPR2018_Feichtenhofer}:
(1) Instead of only predicting a single action label, our method outputs multiple action labels with relevant objects, attributes/relationships and the time of each state transition.
(2) Our action models are learned from semantic-level state transitions based definitions (state detectors are trained on still images), and thus it does not need well-annotated video clips for training. To demonstrate the effectiveness of our action reasoning framework on action recognition, we divide the long video sequences into small clips, each of which contains only one action. The small clips are used to evaluate the performance of our method with \emph{Average Recall} metric to evaluate whether or not the performed actions are recognized.

We compare our method to a representative two-stream (appearance and motion) action recognition algorithm TSN \cite{TPAMI2018_Wang}, which adopts an end-to-end deep learning scheme that utilizes RGB frames and optical flow as a two-stream input. It is worth mentioning that TSN achieves the state-of-the-art performance $94.9\%$ and $89.6\%$ on the benchmark action recognition dataset UCF 101 \cite{CoRR2012_Khurram} and ActivityNet \cite{CVPR2015_Fabian}, respectively. Similar to the other popular action recognition methods \cite{NIPS2014_Simonyan,CVPR2017_Feichtenhofer,ICCV2017_Lee,CVPR2018_Wang,CVPR2018_Feichtenhofer}, the output of TSN is only an action label for a video sequence. 

In our experiment, the TSN model is trained with 50 epochs for both appearance and optical flow models, and the best recognition results are reported for comparison. Let TSN\_RGB be the appearance based model of TSN, TSN\_Flow be the motion based model of TSN, and TSN\_Fusion represent the fused model of appearance and motion. As the results shown in \tab \ref{tbl_tsn_our}, our approach achieves very competitive performance in comparison to different TSN models, especially for ``open'', ``close'', ``drink'' and ``eat'' actions. The average \emph{recall} of TSN\_Flow is $0.71$ and TSN\_Fusion is $0.77$, while the appearance based model TSN\_RGB is only $0.42$. The reason is that in the CAD-120 dataset \cite{IJRR_Hema}, all the 4 subjects (in videos) adopted similar movements to accomplish the same action. Therefore, the motion component contributes disproportionately (w.r.t. appearance ) for the action recognition in these videos. Noted that even without the motion information, the proposed method can still achieve a better recognition performance ($0.80$) comparing to that of the TSN\_Flow (0.71) and TSN\_Fusion (0.77). 

In addition, due to inaccurate video state prediction in some challenging scenes (\eg heavy occlusion), it is difficult to always generate reliable video graph. Therefore, the multiple outputs of our method may contain some false positive actions, and the \emph{Average Precision} is 0.52.

\subsection{Discussion}

In this work, instead of following the traditional action recognition strategy to pursue performance improvement, we would like to advocate a logical reasoning framework for action analysis and understanding in videos. Based on both the logical rules and semantic-level video graph representation, our method enjoys great flexibility and extendability to be applicable for more difficult activity reasoning tasks and other video domains as follows.

{\bf Complex activity reasoning.}
This paper mainly focuses on the atomic action reasoning problem, which can be observed by an attribute or relationship changes in videos. In practice, many complex actions or activities involve a set of objects, attributes and relationships. Similar to the previous work~\cite{CVPR2011_Morariu,CVPR2011_Brendel}  that use prior knowledge and Markov Logic Networks~\cite{Book2009_Russell} for event modeling, our method can also be extended in the same way to reason about complex activities. For example, the ``having\_meal'' activity can be defined as: $\emph{eat} \wedge \emph{drink}$. When both the two atomic actions \emph{eat} and \emph{drink} are detected, the complex activity ``having\_meal'' can be inferred by the first-order logic with predefined rules, as the example shown in the first row of \fig \ref{fig_act_video}.

{\bf Applications on other video domains.}
Since our action reasoning framework is based on prior knowledge and semantic-level video graph representation, it can be easily adapted to a new video domain (such as sports) for explainable action reasoning as long as the knowledge is available. In fact, the prior knowledge used in our framework is commonsense knowledge (as presented in Table 1-4),  which is easy to collect. For other video domains, the only requirement is to replace the prior knowledge (as presented in \tab 1-4) and relevant state detectors.



\section{Conclusion and Future Work}

We proposed an explainable video action reasoning framework based on prior knowledge and semantic-level state transitions. Given a certain video domain, a set of concerned objects, attributes and relationships can be defined based on commonsense knowledge. Moreover, the concerned actions can be explained by the attribute and relationship changes in videos. During the testing stage, given a video, we first generate the scene graph on each video frame for semantic-level state representation, and then construct the video graph by linking the scene graphs by tracking objects across all the video frames sequentially. To this end, our model can detect and explain performed actions by observing state transitions in the video graph. Compared to previous methods, the action reasoning results of our method can be explained by both logical rules and semantic-level video content understanding. Experiments on the daily life dataset show that the proposed method can not only recognizes performed actions, but also provides detailed information to explain how those actions are executed. Moreover, our method can handle multiple concurrent actions in complex videos, just as that of the single action detection. 


In the future, we will construct dataset to empirically study the effectiveness of our framework on complex activity and event detection, as well as its extendability for other domains. In addition, another interesting future work is to automatically learn additional rules by using Probabilistic Inductive Logic Programming~\cite{PILP2008}.

\section*{Acknowledgments}
\small
This research is supported by the National Research Foundation, 
Prime Minister's Office, 
Singapore under its Strategic Capability Research Centres Funding Initiative, 
and the Agency for Science, Technology and Research (A*STAR) under its AME Programmatic Funding Scheme (\#A18A2b0046).
This research is also supported by the NSF of China 61571362, and grant 2018JM6015 of Natural Science Basic Research Plan in Shaanxi Province of China, as well as the Fundamental Research Funds for the Central Universities 3102019ZY1004. Zhiyong Cheng is the corresponding author.

\clearpage

\bibliographystyle{ACM-Reference-Format}
\balance
\bibliography{egbib}


\begin{thebibliography}{53}


\ifx \showCODEN    \undefined \def \showCODEN     #1{\unskip}     \fi
\ifx \showDOI      \undefined \def \showDOI       #1{#1}\fi
\ifx \showISBNx    \undefined \def \showISBNx     #1{\unskip}     \fi
\ifx \showISBNxiii \undefined \def \showISBNxiii  #1{\unskip}     \fi
\ifx \showISSN     \undefined \def \showISSN      #1{\unskip}     \fi
\ifx \showLCCN     \undefined \def \showLCCN      #1{\unskip}     \fi
\ifx \shownote     \undefined \def \shownote      #1{#1}          \fi
\ifx \showarticletitle \undefined \def \showarticletitle #1{#1}   \fi
\ifx \showURL      \undefined \def \showURL       {\relax}        \fi
\providecommand\bibfield[2]{#2}
\providecommand\bibinfo[2]{#2}
\providecommand\natexlab[1]{#1}
\providecommand\showeprint[2][]{arXiv:#2}

\bibitem[\protect\citeauthoryear{Alayrac, Sivic, Laptev, and
  Lacoste-Julien}{Alayrac et~al\mbox{.}}{2017}]%
        {ICCV2017_Alayrac}
\bibfield{author}{\bibinfo{person}{Jean-Baptiste Alayrac},
  \bibinfo{person}{Josef Sivic}, \bibinfo{person}{Ivan Laptev}, {and}
  \bibinfo{person}{Simon Lacoste-Julien}.} \bibinfo{year}{2017}\natexlab{}.
\newblock \showarticletitle{Joint discovery of object states and manipulating
  actions}. In \bibinfo{booktitle}{\emph{ICCV}}. IEEE.
\newblock


\bibitem[\protect\citeauthoryear{Brendel, Fern, and Todorovic}{Brendel
  et~al\mbox{.}}{2011}]%
        {CVPR2011_Brendel}
\bibfield{author}{\bibinfo{person}{William Brendel}, \bibinfo{person}{Alan
  Fern}, {and} \bibinfo{person}{Sinisa Todorovic}.}
  \bibinfo{year}{2011}\natexlab{}.
\newblock \showarticletitle{Probabilistic event logic for interval-based event
  recognition}. In \bibinfo{booktitle}{\emph{CVPR}}. IEEE.
\newblock


\bibitem[\protect\citeauthoryear{Carreira and Zisserman}{Carreira and
  Zisserman}{2017}]%
        {CVPR2017_Carreira}
\bibfield{author}{\bibinfo{person}{Joao Carreira} {and} \bibinfo{person}{Andrew
  Zisserman}.} \bibinfo{year}{2017}\natexlab{}.
\newblock \showarticletitle{Quo vadis, action recognition? a new model and the
  kinetics dataset}. In \bibinfo{booktitle}{\emph{CVPR}}. IEEE.
\newblock


\bibitem[\protect\citeauthoryear{Dai, Zhang, and Lin}{Dai
  et~al\mbox{.}}{2017}]%
        {CVPR2017_Dai}
\bibfield{author}{\bibinfo{person}{Bo Dai}, \bibinfo{person}{Yuqi Zhang}, {and}
  \bibinfo{person}{Dahua Lin}.} \bibinfo{year}{2017}\natexlab{}.
\newblock \showarticletitle{Detecting Visual Relationships With Deep Relational
  Networks}. In \bibinfo{booktitle}{\emph{CVPR}}. IEEE.
\newblock


\bibitem[\protect\citeauthoryear{De~Raedt and Kersting}{De~Raedt and
  Kersting}{2008}]%
        {PILP2008}
\bibfield{author}{\bibinfo{person}{Luc De~Raedt} {and}
  \bibinfo{person}{Kristian Kersting}.} \bibinfo{year}{2008}\natexlab{}.
\newblock \showarticletitle{Probabilistic inductive logic programming}.
\newblock In \bibinfo{booktitle}{\emph{Probabilistic Inductive Logic
  Programming}}. \bibinfo{publisher}{Springer}, \bibinfo{pages}{1--27}.
\newblock


\bibitem[\protect\citeauthoryear{Fabian Caba~Heilbron and Niebles}{Fabian
  Caba~Heilbron and Niebles}{2015}]%
        {CVPR2015_Fabian}
\bibfield{author}{\bibinfo{person}{Bernard~Ghanem Fabian Caba~Heilbron,
  Victor~Escorcia} {and} \bibinfo{person}{Juan~Carlos Niebles}.}
  \bibinfo{year}{2015}\natexlab{}.
\newblock \showarticletitle{ActivityNet: A Large-Scale Video Benchmark for
  Human Activity Understanding}. In \bibinfo{booktitle}{\emph{CVPR}}. IEEE.
\newblock


\bibitem[\protect\citeauthoryear{Fathi and Rehg}{Fathi and Rehg}{2013}]%
        {CVPR2013_Fathi}
\bibfield{author}{\bibinfo{person}{Alireza Fathi} {and}
  \bibinfo{person}{James~M. Rehg}.} \bibinfo{year}{2013}\natexlab{}.
\newblock \showarticletitle{Modeling Actions through State Changes}. In
  \bibinfo{booktitle}{\emph{CVPR}}. IEEE.
\newblock


\bibitem[\protect\citeauthoryear{Feichtenhofer, Pinz, and Wildes}{Feichtenhofer
  et~al\mbox{.}}{2017}]%
        {CVPR2017_Feichtenhofer}
\bibfield{author}{\bibinfo{person}{Christoph Feichtenhofer},
  \bibinfo{person}{Axel Pinz}, {and} \bibinfo{person}{Richard~P Wildes}.}
  \bibinfo{year}{2017}\natexlab{}.
\newblock \showarticletitle{Spatiotemporal multiplier networks for video action
  recognition}. In \bibinfo{booktitle}{\emph{CVPR}}. IEEE.
\newblock


\bibitem[\protect\citeauthoryear{Feichtenhofer, Pinz, Wildes, and
  Zisserman}{Feichtenhofer et~al\mbox{.}}{2018}]%
        {CVPR2018_Feichtenhofer}
\bibfield{author}{\bibinfo{person}{Christoph Feichtenhofer},
  \bibinfo{person}{Axel Pinz}, \bibinfo{person}{Richard~P. Wildes}, {and}
  \bibinfo{person}{Andrew Zisserman}.} \bibinfo{year}{2018}\natexlab{}.
\newblock \showarticletitle{What Have We Learned From Deep Representations for
  Action Recognition?}. In \bibinfo{booktitle}{\emph{CVPR}}. IEEE.
\newblock


\bibitem[\protect\citeauthoryear{Fire and Zhu}{Fire and Zhu}{2015}]%
        {TIST_Fire2015}
\bibfield{author}{\bibinfo{person}{Amy Fire} {and} \bibinfo{person}{Song-Chun
  Zhu}.} \bibinfo{year}{2015}\natexlab{}.
\newblock \showarticletitle{Learning Perceptual Causality from Video}.
\newblock \bibinfo{journal}{\emph{Transactions on Intelligent Systems and
  Technology}} \bibinfo{volume}{7}, \bibinfo{number}{2} (\bibinfo{year}{2015}),
  \bibinfo{pages}{23:1--23:22}.
\newblock


\bibitem[\protect\citeauthoryear{Graves}{Graves}{2012}]%
        {SSL2012_Graves}
\bibfield{author}{\bibinfo{person}{Alex Graves}.}
  \bibinfo{year}{2012}\natexlab{}.
\newblock \showarticletitle{Supervised sequence labelling}.
\newblock In \bibinfo{booktitle}{\emph{Supervised sequence labelling with
  recurrent neural networks}}. \bibinfo{publisher}{Springer},
  \bibinfo{pages}{5--13}.
\newblock


\bibitem[\protect\citeauthoryear{Guo, Chou, Huang, Song, Yeung, and
  Fei-Fei}{Guo et~al\mbox{.}}{2018}]%
        {ECCV2018_Guo}
\bibfield{author}{\bibinfo{person}{Michelle Guo}, \bibinfo{person}{Edward
  Chou}, \bibinfo{person}{De-An Huang}, \bibinfo{person}{Shuran Song},
  \bibinfo{person}{Serena Yeung}, {and} \bibinfo{person}{Li Fei-Fei}.}
  \bibinfo{year}{2018}\natexlab{}.
\newblock \showarticletitle{Neural graph matching networks for fewshot 3d
  action recognition}. In \bibinfo{booktitle}{\emph{ECCV}}. Springer.
\newblock


\bibitem[\protect\citeauthoryear{He, Zhang, Ren, and Sun}{He
  et~al\mbox{.}}{2016}]%
        {CVPR2016_He}
\bibfield{author}{\bibinfo{person}{Kaiming He}, \bibinfo{person}{Xiangyu
  Zhang}, \bibinfo{person}{Shaoqing Ren}, {and} \bibinfo{person}{Jian Sun}.}
  \bibinfo{year}{2016}\natexlab{}.
\newblock \showarticletitle{Deep residual learning for image recognition}. In
  \bibinfo{booktitle}{\emph{CVPR}}. IEEE.
\newblock


\bibitem[\protect\citeauthoryear{Ijsselmuiden and Stiefelhagen}{Ijsselmuiden
  and Stiefelhagen}{2010}]%
        {AAAI2010_Ijsselmuiden}
\bibfield{author}{\bibinfo{person}{Joris Ijsselmuiden} {and}
  \bibinfo{person}{Rainer Stiefelhagen}.} \bibinfo{year}{2010}\natexlab{}.
\newblock \showarticletitle{Towards high-level human activity recognition
  through computer vision and temporal logic}. In
  \bibinfo{booktitle}{\emph{AAAI}}.
\newblock


\bibitem[\protect\citeauthoryear{Intille and Bobick}{Intille and
  Bobick}{2001}]%
        {CVIU2001_Intille}
\bibfield{author}{\bibinfo{person}{Stephen~S Intille} {and}
  \bibinfo{person}{Aaron~F Bobick}.} \bibinfo{year}{2001}\natexlab{}.
\newblock \showarticletitle{Recognizing planned, multiperson action}.
\newblock \bibinfo{journal}{\emph{Computer Vision and Image Understanding
  (CVIU)}} \bibinfo{volume}{81}, \bibinfo{number}{3} (\bibinfo{year}{2001}),
  \bibinfo{pages}{414--445}.
\newblock


\bibitem[\protect\citeauthoryear{Jain, Zamir, Savarese, and Saxena}{Jain
  et~al\mbox{.}}{2016}]%
        {CVPR2016_Jain}
\bibfield{author}{\bibinfo{person}{Ashesh Jain}, \bibinfo{person}{Amir~R
  Zamir}, \bibinfo{person}{Silvio Savarese}, {and} \bibinfo{person}{Ashutosh
  Saxena}.} \bibinfo{year}{2016}\natexlab{}.
\newblock \showarticletitle{Structural-RNN: Deep learning on spatio-temporal
  graphs}. In \bibinfo{booktitle}{\emph{CVPR}}. IEEE,
  \bibinfo{pages}{5308--5317}.
\newblock


\bibitem[\protect\citeauthoryear{Kingma and Ba}{Kingma and Ba}{2014}]%
        {CoRR2014_Adam}
\bibfield{author}{\bibinfo{person}{Diederik~P. Kingma} {and}
  \bibinfo{person}{Jimmy Ba}.} \bibinfo{year}{2014}\natexlab{}.
\newblock \showarticletitle{Adam: {A} Method for Stochastic Optimization}.
\newblock \bibinfo{journal}{\emph{CoRR}}  \bibinfo{volume}{abs/1412.6980}
  (\bibinfo{year}{2014}).
\newblock


\bibitem[\protect\citeauthoryear{Kipf and Welling}{Kipf and Welling}{2017}]%
        {ICLR2017_Kipf}
\bibfield{author}{\bibinfo{person}{Thomas~N Kipf} {and} \bibinfo{person}{Max
  Welling}.} \bibinfo{year}{2017}\natexlab{}.
\newblock \showarticletitle{Semi-supervised classification with graph
  convolutional networks}. In \bibinfo{booktitle}{\emph{ICLR}}.
\newblock


\bibitem[\protect\citeauthoryear{Klaser, Marszalek, and Schmid}{Klaser
  et~al\mbox{.}}{2008}]%
        {BMVC2008_Klaser}
\bibfield{author}{\bibinfo{person}{Alexander Klaser}, \bibinfo{person}{Marcin
  Marszalek}, {and} \bibinfo{person}{Cordelia Schmid}.}
  \bibinfo{year}{2008}\natexlab{}.
\newblock \showarticletitle{A Spatio-Temporal Descriptor Based on
  3D-Gradients}. In \bibinfo{booktitle}{\emph{BMVC}}.
\newblock
\urldef\tempurl%
\url{https://hal.inria.fr/inria-00514853}
\showURL{%
\tempurl}


\bibitem[\protect\citeauthoryear{Koppula, Gupta, and Saxena}{Koppula
  et~al\mbox{.}}{2013}]%
        {IJRR_Hema}
\bibfield{author}{\bibinfo{person}{Hema~Swetha Koppula},
  \bibinfo{person}{Rudhir Gupta}, {and} \bibinfo{person}{Ashutosh Saxena}.}
  \bibinfo{year}{2013}\natexlab{}.
\newblock \showarticletitle{Learning human activities and object affordances
  from RGB-D videos}.
\newblock \bibinfo{journal}{\emph{The International Journal of Robotics
  Research}} \bibinfo{volume}{32}, \bibinfo{number}{8} (\bibinfo{year}{2013}),
  \bibinfo{pages}{951--970}.
\newblock


\bibitem[\protect\citeauthoryear{Krishna, Chami, Bernstein, and
  Fei-Fei}{Krishna et~al\mbox{.}}{2018}]%
        {CVPR2018_Krishna}
\bibfield{author}{\bibinfo{person}{Ranjay Krishna}, \bibinfo{person}{Ines
  Chami}, \bibinfo{person}{Michael Bernstein}, {and} \bibinfo{person}{Li
  Fei-Fei}.} \bibinfo{year}{2018}\natexlab{}.
\newblock \showarticletitle{Referring Relationships}. In
  \bibinfo{booktitle}{\emph{CVPR}}. IEEE.
\newblock


\bibitem[\protect\citeauthoryear{Krishna, Zhu, Groth, Johnson, Hata, Kravitz,
  Chen, Kalantidis, Li, Shamma, Bernstein, and Fei-Fei}{Krishna
  et~al\mbox{.}}{2017}]%
        {IJCV2017_Krishna}
\bibfield{author}{\bibinfo{person}{Ranjay Krishna}, \bibinfo{person}{Yuke Zhu},
  \bibinfo{person}{Oliver Groth}, \bibinfo{person}{Justin Johnson},
  \bibinfo{person}{Kenji Hata}, \bibinfo{person}{Joshua Kravitz},
  \bibinfo{person}{Stephanie Chen}, \bibinfo{person}{Yannis Kalantidis},
  \bibinfo{person}{Li-Jia Li}, \bibinfo{person}{David~A. Shamma},
  \bibinfo{person}{Michael~S. Bernstein}, {and} \bibinfo{person}{Li Fei-Fei}.}
  \bibinfo{year}{2017}\natexlab{}.
\newblock \showarticletitle{Visual Genome: Connecting Language and Vision Using
  Crowdsourced Dense Image Annotations}.
\newblock \bibinfo{journal}{\emph{International Journal of Computer Vision
  (IJCV)}} \bibinfo{volume}{123}, \bibinfo{number}{1} (\bibinfo{year}{2017}),
  \bibinfo{pages}{32--73}.
\newblock


\bibitem[\protect\citeauthoryear{Lee, Kim, Kang, and Lee}{Lee
  et~al\mbox{.}}{2017}]%
        {ICCV2017_Lee}
\bibfield{author}{\bibinfo{person}{Inwoong Lee}, \bibinfo{person}{Doyoung Kim},
  \bibinfo{person}{Seoungyoon Kang}, {and} \bibinfo{person}{Sanghoon Lee}.}
  \bibinfo{year}{2017}\natexlab{}.
\newblock \showarticletitle{Ensemble deep learning for skeleton-based action
  recognition using temporal sliding LSTM networks}. In
  \bibinfo{booktitle}{\emph{ICCV}}. IEEE.
\newblock


\bibitem[\protect\citeauthoryear{Liu, Su, Nie, and Kankanhalli}{Liu
  et~al\mbox{.}}{2017a}]%
        {TPAMI2017_Liu}
\bibfield{author}{\bibinfo{person}{An-An Liu}, \bibinfo{person}{Yu-Ting Su},
  \bibinfo{person}{Wei-Zhi Nie}, {and} \bibinfo{person}{Mohan Kankanhalli}.}
  \bibinfo{year}{2017}\natexlab{a}.
\newblock \showarticletitle{Hierarchical clustering multi-task learning for
  joint human action grouping and recognition}.
\newblock \bibinfo{journal}{\emph{IEEE transactions on pattern analysis and
  machine intelligence (TPAMI)}} \bibinfo{volume}{39}, \bibinfo{number}{1}
  (\bibinfo{year}{2017}), \bibinfo{pages}{102--114}.
\newblock


\bibitem[\protect\citeauthoryear{Liu, Wei, and Zhu}{Liu et~al\mbox{.}}{2017b}]%
        {ICCV2017_Liu}
\bibfield{author}{\bibinfo{person}{Yang Liu}, \bibinfo{person}{Ping Wei}, {and}
  \bibinfo{person}{Song-Chun Zhu}.} \bibinfo{year}{2017}\natexlab{b}.
\newblock \showarticletitle{Jointly Recognizing Object Fluents and Tasks in
  Egocentric Videos}. In \bibinfo{booktitle}{\emph{ICCV}}. IEEE.
\newblock


\bibitem[\protect\citeauthoryear{Lu, Krishna, Bernstein, and Fei-Fei}{Lu
  et~al\mbox{.}}{2016}]%
        {ECCV2016_Lu}
\bibfield{author}{\bibinfo{person}{Cewu Lu}, \bibinfo{person}{Ranjay Krishna},
  \bibinfo{person}{Michael Bernstein}, {and} \bibinfo{person}{Li Fei-Fei}.}
  \bibinfo{year}{2016}\natexlab{}.
\newblock \showarticletitle{Visual relationship detection with language
  priors}. In \bibinfo{booktitle}{\emph{ECCV}}. Springer.
\newblock


\bibitem[\protect\citeauthoryear{Lukezic, Vojir, ˇCehovin~Zajc, Matas, and
  Kristan}{Lukezic et~al\mbox{.}}{2017}]%
        {CVPR2017_Lukezic}
\bibfield{author}{\bibinfo{person}{Alan Lukezic}, \bibinfo{person}{Tomas
  Vojir}, \bibinfo{person}{Luka ˇCehovin~Zajc}, \bibinfo{person}{Jiri Matas},
  {and} \bibinfo{person}{Matej Kristan}.} \bibinfo{year}{2017}\natexlab{}.
\newblock \showarticletitle{Discriminative correlation filter with channel and
  spatial reliability}. In \bibinfo{booktitle}{\emph{CVPR}}. IEEE.
\newblock


\bibitem[\protect\citeauthoryear{Morariu and Davis}{Morariu and Davis}{2011}]%
        {CVPR2011_Morariu}
\bibfield{author}{\bibinfo{person}{Vlad~I Morariu} {and}
  \bibinfo{person}{Larry~S Davis}.} \bibinfo{year}{2011}\natexlab{}.
\newblock \showarticletitle{Multi-agent event recognition in structured
  scenarios}. In \bibinfo{booktitle}{\emph{CVPR}}. IEEE.
\newblock


\bibitem[\protect\citeauthoryear{Pednault}{Pednault}{1994}]%
        {JLC1994_Pednault}
\bibfield{author}{\bibinfo{person}{Edwin~PD Pednault}.}
  \bibinfo{year}{1994}\natexlab{}.
\newblock \showarticletitle{ADL and the state-transition model of action}.
\newblock \bibinfo{journal}{\emph{Journal of logic and computation}}
  \bibinfo{volume}{4}, \bibinfo{number}{5} (\bibinfo{year}{1994}),
  \bibinfo{pages}{467--512}.
\newblock


\bibitem[\protect\citeauthoryear{Poole and Mackworth}{Poole and
  Mackworth}{2010}]%
        {Book2010_Poole}
\bibfield{author}{\bibinfo{person}{David~L Poole} {and} \bibinfo{person}{Alan~K
  Mackworth}.} \bibinfo{year}{2010}\natexlab{}.
\newblock \bibinfo{booktitle}{\emph{Artificial Intelligence: foundations of
  computational agents}}.
\newblock \bibinfo{publisher}{Cambridge University Press}.
\newblock


\bibitem[\protect\citeauthoryear{Qi, Wang, Jia, Shen, and Zhu}{Qi
  et~al\mbox{.}}{2018}]%
        {ECCV2018_Qi}
\bibfield{author}{\bibinfo{person}{Siyuan Qi}, \bibinfo{person}{Wenguan Wang},
  \bibinfo{person}{Baoxiong Jia}, \bibinfo{person}{Jianbing Shen}, {and}
  \bibinfo{person}{Song-Chun Zhu}.} \bibinfo{year}{2018}\natexlab{}.
\newblock \showarticletitle{Learning human-object interactions by graph parsing
  neural networks}. In \bibinfo{booktitle}{\emph{ECCV}}. Springer,
  \bibinfo{pages}{401--417}.
\newblock


\bibitem[\protect\citeauthoryear{Richardson and Domingos}{Richardson and
  Domingos}{2006}]%
        {ML2006_Richardson}
\bibfield{author}{\bibinfo{person}{Matthew Richardson} {and}
  \bibinfo{person}{Pedro Domingos}.} \bibinfo{year}{2006}\natexlab{}.
\newblock \showarticletitle{Markov logic networks}.
\newblock \bibinfo{journal}{\emph{Machine learning}} \bibinfo{volume}{62},
  \bibinfo{number}{1-2} (\bibinfo{year}{2006}), \bibinfo{pages}{107--136}.
\newblock


\bibitem[\protect\citeauthoryear{Russell and Norvig}{Russell and
  Norvig}{2009}]%
        {Book2009_Russell}
\bibfield{author}{\bibinfo{person}{Stuart Russell} {and} \bibinfo{person}{Peter
  Norvig}.} \bibinfo{year}{2009}\natexlab{}.
\newblock \bibinfo{booktitle}{\emph{Artificial Intelligence: A Modern Approach}
  (\bibinfo{edition}{3rd} ed.)}.
\newblock \bibinfo{publisher}{Prentice Hall Press}.
\newblock


\bibitem[\protect\citeauthoryear{Schmid, Rozenfeld, Marszalek, and
  Laptev}{Schmid et~al\mbox{.}}{2008}]%
        {CVPR2008_Cordelia}
\bibfield{author}{\bibinfo{person}{Cordelia Schmid}, \bibinfo{person}{Benjamin
  Rozenfeld}, \bibinfo{person}{Marcin Marszalek}, {and} \bibinfo{person}{Ivan
  Laptev}.} \bibinfo{year}{2008}\natexlab{}.
\newblock \showarticletitle{Learning realistic human actions from movies}. In
  \bibinfo{booktitle}{\emph{CVPR}}. IEEE.
\newblock


\bibitem[\protect\citeauthoryear{Shotton, Fitzgibbon, Cook, Sharp, Finocchio,
  Moore, Kipman, and Blake}{Shotton et~al\mbox{.}}{2011}]%
        {CVPR2011_Shotton}
\bibfield{author}{\bibinfo{person}{J. Shotton}, \bibinfo{person}{A.
  Fitzgibbon}, \bibinfo{person}{M. Cook}, \bibinfo{person}{T. Sharp},
  \bibinfo{person}{M. Finocchio}, \bibinfo{person}{R. Moore},
  \bibinfo{person}{A. Kipman}, {and} \bibinfo{person}{A. Blake}.}
  \bibinfo{year}{2011}\natexlab{}.
\newblock \showarticletitle{Real-time Human Pose Recognition in Parts from
  Single Depth Images}. In \bibinfo{booktitle}{\emph{CVPR}}. IEEE.
\newblock


\bibitem[\protect\citeauthoryear{Simonyan and Zisserman}{Simonyan and
  Zisserman}{2014a}]%
        {NIPS2014_Simonyan}
\bibfield{author}{\bibinfo{person}{Karen Simonyan} {and}
  \bibinfo{person}{Andrew Zisserman}.} \bibinfo{year}{2014}\natexlab{a}.
\newblock \showarticletitle{Two-Stream Convolutional Networks for Action
  Recognition in Videos}.
\newblock In \bibinfo{booktitle}{\emph{NeurIPS}}. \bibinfo{pages}{568--576}.
\newblock


\bibitem[\protect\citeauthoryear{Simonyan and Zisserman}{Simonyan and
  Zisserman}{2014b}]%
        {CoRR2014_Karen}
\bibfield{author}{\bibinfo{person}{Karen Simonyan} {and}
  \bibinfo{person}{Andrew Zisserman}.} \bibinfo{year}{2014}\natexlab{b}.
\newblock \showarticletitle{Very Deep Convolutional Networks for Large-Scale
  Image Recognition}.
\newblock \bibinfo{journal}{\emph{CoRR}}  \bibinfo{volume}{abs/1409.1556}
  (\bibinfo{year}{2014}).
\newblock
\urldef\tempurl%
\url{http://arxiv.org/abs/1409.1556}
\showURL{%
\tempurl}


\bibitem[\protect\citeauthoryear{Soomro, Zamir, and Shah}{Soomro
  et~al\mbox{.}}{2012}]%
        {CoRR2012_Khurram}
\bibfield{author}{\bibinfo{person}{Khurram Soomro},
  \bibinfo{person}{Amir~Roshan Zamir}, {and} \bibinfo{person}{Mubarak Shah}.}
  \bibinfo{year}{2012}\natexlab{}.
\newblock \showarticletitle{UCF101: A Dataset of 101 Human Actions Classes From
  Videos in The Wild}.
\newblock \bibinfo{journal}{\emph{CoRR}}  \bibinfo{volume}{abs/1212.0402}
  (\bibinfo{year}{2012}).
\newblock


\bibitem[\protect\citeauthoryear{Sowa}{Sowa}{2014}]%
        {Book2014_Sowa}
\bibfield{author}{\bibinfo{person}{John~F Sowa}.}
  \bibinfo{year}{2014}\natexlab{}.
\newblock \bibinfo{booktitle}{\emph{Principles of semantic networks:
  Explorations in the representation of knowledge}}.
\newblock \bibinfo{publisher}{Morgan Kaufmann}.
\newblock


\bibitem[\protect\citeauthoryear{Szegedy, Ioffe, Vanhoucke, and Alemi}{Szegedy
  et~al\mbox{.}}{2017}]%
        {AAAI2017_Szegedy}
\bibfield{author}{\bibinfo{person}{Christian Szegedy}, \bibinfo{person}{Sergey
  Ioffe}, \bibinfo{person}{Vincent Vanhoucke}, {and}
  \bibinfo{person}{Alexander~A Alemi}.} \bibinfo{year}{2017}\natexlab{}.
\newblock \showarticletitle{Inception-v4, inception-resnet and the impact of
  residual connections on learning}. In \bibinfo{booktitle}{\emph{AAAI}}.
\newblock


\bibitem[\protect\citeauthoryear{Tran, Wang, Torresani, Ray, LeCun, and
  Paluri}{Tran et~al\mbox{.}}{2018}]%
        {CVPR2018_Tran}
\bibfield{author}{\bibinfo{person}{Du Tran}, \bibinfo{person}{Heng Wang},
  \bibinfo{person}{Lorenzo Torresani}, \bibinfo{person}{Jamie Ray},
  \bibinfo{person}{Yann LeCun}, {and} \bibinfo{person}{Manohar Paluri}.}
  \bibinfo{year}{2018}\natexlab{}.
\newblock \showarticletitle{A closer look at spatiotemporal convolutions for
  action recognition}. In \bibinfo{booktitle}{\emph{CVPR}}. IEEE.
\newblock


\bibitem[\protect\citeauthoryear{Tran and Davis}{Tran and Davis}{2008}]%
        {ECCV2008_Tran}
\bibfield{author}{\bibinfo{person}{Son~D Tran} {and} \bibinfo{person}{Larry~S
  Davis}.} \bibinfo{year}{2008}\natexlab{}.
\newblock \showarticletitle{Event modeling and recognition using markov logic
  networks}. In \bibinfo{booktitle}{\emph{ECCV}}. Springer.
\newblock


\bibitem[\protect\citeauthoryear{Wang and Schmid}{Wang and Schmid}{2013}]%
        {ICCV2013_Wang}
\bibfield{author}{\bibinfo{person}{Heng Wang} {and} \bibinfo{person}{Cordelia
  Schmid}.} \bibinfo{year}{2013}\natexlab{}.
\newblock \showarticletitle{Action Recognition with Improved Trajectories}. In
  \bibinfo{booktitle}{\emph{ICCV}}. IEEE.
\newblock


\bibitem[\protect\citeauthoryear{Wang, Li, Li, and Van~Gool}{Wang
  et~al\mbox{.}}{2018a}]%
        {CVPR2018_Wang}
\bibfield{author}{\bibinfo{person}{Limin Wang}, \bibinfo{person}{Wei Li},
  \bibinfo{person}{Wen Li}, {and} \bibinfo{person}{Luc Van~Gool}.}
  \bibinfo{year}{2018}\natexlab{a}.
\newblock \showarticletitle{Appearance-and-Relation Networks for Video
  Classification}. In \bibinfo{booktitle}{\emph{CVPR}}. IEEE.
\newblock


\bibitem[\protect\citeauthoryear{Wang, Xiong, Wang, Qiao, Lin, Tang, and
  Van~Gool}{Wang et~al\mbox{.}}{2018c}]%
        {TPAMI2018_Wang}
\bibfield{author}{\bibinfo{person}{Limin Wang}, \bibinfo{person}{Yuanjun
  Xiong}, \bibinfo{person}{Zhe Wang}, \bibinfo{person}{Yu Qiao},
  \bibinfo{person}{Dahua Lin}, \bibinfo{person}{Xiaoou Tang}, {and}
  \bibinfo{person}{Luc Van~Gool}.} \bibinfo{year}{2018}\natexlab{c}.
\newblock \showarticletitle{Temporal segment networks for action recognition in
  videos}.
\newblock \bibinfo{journal}{\emph{IEEE transactions on pattern analysis and
  machine intelligence (TPAMI)}} (\bibinfo{year}{2018}).
\newblock


\bibitem[\protect\citeauthoryear{Wang, Hong, Li, Zha, Yan, and Chua}{Wang
  et~al\mbox{.}}{2012a}]%
        {TMM2012_Event}
\bibfield{author}{\bibinfo{person}{Meng Wang}, \bibinfo{person}{Richang Hong},
  \bibinfo{person}{Guangda Li}, \bibinfo{person}{Zheng{-}Jun Zha},
  \bibinfo{person}{Shuicheng Yan}, {and} \bibinfo{person}{Tat{-}Seng Chua}.}
  \bibinfo{year}{2012}\natexlab{a}.
\newblock \showarticletitle{Event Driven Web Video Summarization by Tag
  Localization and Key-Shot Identification}.
\newblock \bibinfo{journal}{\emph{IEEE Transactions on Multimedia (TMM)}}
  \bibinfo{volume}{14}, \bibinfo{number}{4} (\bibinfo{year}{2012}),
  \bibinfo{pages}{975--985}.
\newblock


\bibitem[\protect\citeauthoryear{Wang, Hong, Yuan, Yan, and Chua}{Wang
  et~al\mbox{.}}{2012b}]%
        {TMM2012_Movie}
\bibfield{author}{\bibinfo{person}{Meng Wang}, \bibinfo{person}{Richang Hong},
  \bibinfo{person}{Xiao{-}Tong Yuan}, \bibinfo{person}{Shuicheng Yan}, {and}
  \bibinfo{person}{Tat{-}Seng Chua}.} \bibinfo{year}{2012}\natexlab{b}.
\newblock \showarticletitle{Movie2Comics: Towards a Lively Video Content
  Presentation}.
\newblock \bibinfo{journal}{\emph{IEEE Transactions on Multimedia (TMM)}}
  \bibinfo{volume}{14}, \bibinfo{number}{3-2} (\bibinfo{year}{2012}),
  \bibinfo{pages}{858--870}.
\newblock


\bibitem[\protect\citeauthoryear{Wang, Luo, Ni, Yuan, Wang, and Yan}{Wang
  et~al\mbox{.}}{2018b}]%
        {TCSVT2018_Wang}
\bibfield{author}{\bibinfo{person}{Meng Wang}, \bibinfo{person}{Changzhi Luo},
  \bibinfo{person}{Bingbing Ni}, \bibinfo{person}{Jun Yuan},
  \bibinfo{person}{Jianfeng Wang}, {and} \bibinfo{person}{Shuicheng Yan}.}
  \bibinfo{year}{2018}\natexlab{b}.
\newblock \showarticletitle{First-Person Daily Activity Recognition With
  Manipulated Object Proposals and Non-Linear Feature Fusion}.
\newblock \bibinfo{journal}{\emph{IEEE Transactions on Circuits and Systems for
  Video Technology (TCSVT)}} \bibinfo{volume}{28}, \bibinfo{number}{10}
  (\bibinfo{year}{2018}), \bibinfo{pages}{2946--2955}.
\newblock


\bibitem[\protect\citeauthoryear{Wang, Farhadi, and Gupta}{Wang
  et~al\mbox{.}}{2016}]%
        {CVPR2016_Wang}
\bibfield{author}{\bibinfo{person}{Xiaolong Wang}, \bibinfo{person}{Ali
  Farhadi}, {and} \bibinfo{person}{Abhinav Gupta}.}
  \bibinfo{year}{2016}\natexlab{}.
\newblock \showarticletitle{Actions ~ Transformations}. In
  \bibinfo{booktitle}{\emph{CVPR}}. IEEE.
\newblock


\bibitem[\protect\citeauthoryear{Wang and Gupta}{Wang and Gupta}{2018}]%
        {ECCV2018_Wang}
\bibfield{author}{\bibinfo{person}{Xiaolong Wang} {and}
  \bibinfo{person}{Abhinav Gupta}.} \bibinfo{year}{2018}\natexlab{}.
\newblock \showarticletitle{Videos as space-time region graphs}. In
  \bibinfo{booktitle}{\emph{ECCV}}. Springer.
\newblock


\bibitem[\protect\citeauthoryear{Xu, Zhu, Choy, and Fei-Fei}{Xu
  et~al\mbox{.}}{2017}]%
        {CVPR2017_Xu}
\bibfield{author}{\bibinfo{person}{Danfei Xu}, \bibinfo{person}{Yuke Zhu},
  \bibinfo{person}{Christopher~B. Choy}, {and} \bibinfo{person}{Li Fei-Fei}.}
  \bibinfo{year}{2017}\natexlab{}.
\newblock \showarticletitle{Scene Graph Generation by Iterative Message
  Passing}. In \bibinfo{booktitle}{\emph{CVPR}}. IEEE.
\newblock


\bibitem[\protect\citeauthoryear{Xu, Liu, Wong, Zhang, Nie, Su, and
  Kankanhalli}{Xu et~al\mbox{.}}{2018}]%
        {TCSVT2018_Xu}
\bibfield{author}{\bibinfo{person}{Ning Xu}, \bibinfo{person}{An-An Liu},
  \bibinfo{person}{Yongkang Wong}, \bibinfo{person}{Yongdong Zhang},
  \bibinfo{person}{Weizhi Nie}, \bibinfo{person}{Yuting Su}, {and}
  \bibinfo{person}{Mohan Kankanhalli}.} \bibinfo{year}{2018}\natexlab{}.
\newblock \showarticletitle{Dual-stream recurrent neural network for video
  captioning}.
\newblock \bibinfo{journal}{\emph{IEEE Transactions on Circuits and Systems for
  Video Technology (TCSVT)}} (\bibinfo{year}{2018}).
\newblock


\bibitem[\protect\citeauthoryear{Zhu, Gordon, Kolve, Fox, Fei-Fei, Gupta,
  Mottaghi, and Farhadi}{Zhu et~al\mbox{.}}{2017}]%
        {ICCV2017_Zhu}
\bibfield{author}{\bibinfo{person}{Yuke Zhu}, \bibinfo{person}{Daniel Gordon},
  \bibinfo{person}{Eric Kolve}, \bibinfo{person}{Dieter Fox},
  \bibinfo{person}{Li Fei-Fei}, \bibinfo{person}{Abhinav Gupta},
  \bibinfo{person}{Roozbeh Mottaghi}, {and} \bibinfo{person}{Ali Farhadi}.}
  \bibinfo{year}{2017}\natexlab{}.
\newblock \showarticletitle{Visual semantic planning using deep successor
  representations}. In \bibinfo{booktitle}{\emph{ICCV}}. IEEE.
\newblock


\end{thebibliography}

\end{document}